# Enhancing Plagiarism Detection in Marathi with a Weighted Ensemble of TF-IDF and BERT Embeddings for Low-Resource Language Processing

**Atharva Mutsaddi and Aditya Choudhary**
Department of Computer Science and Engineering, COEP Technological University
`atharvaam21.comp@coeptech.ac.in, choudharyap21.comp@coeptech.ac.in`

## Abstract

Plagiarism involves using another persons work or concepts without proper attribution, presenting them as original creations. With the growing amount of data communicated in regional languages such as Marathi - one of India's regional languages - it is crucial to design robust plagiarism detection systems tailored for low-resource languages. Language models like Bidirectional Encoder Representations from Transformers (BERT) have demonstrated exceptional capability in text representation and feature extraction, making them essential tools for semantic analysis and plagiarism detection. However, the application of BERT for low-resource languages remains under-explored, particularly in the context of plagiarism detection. This paper presents a method to enhance the accuracy of plagiarism detection for Marathi texts using BERT sentence embeddings in conjunction with Term Frequency-Inverse Document Frequency (TF-IDF) feature representation. This approach effectively captures statistical, semantic, and syntactic aspects of text features through a weighted voting ensemble of machine learning models.

## 1 Introduction

Plagiarism is a pervasive issue across various industries. While extensive research has focused on detecting plagiarized texts in widely spoken languages like English, similar advancements for regional languages, particularly Marathi - a language spoken in India - are lacking. Language models for text representation, such as BERT (Devlin et al., 2018), which are often used for semantic-based plagiarism detection, are significantly more robust for these commonly spoken languages due to the abundance of training corpora. In contrast, the scarcity of resources for Marathi leads to weaker semantic analysis, resulting in less accurate plagiarism detection.

Most existing approaches to plagiarism detection in Marathi rely on techniques such as syntax, fuzzy matching, structural analysis, or stylometry (Kulkarni et al., 2021), which often overlook the meaning of texts and the linguistic nuances involved. Consequently, these methods can yield inaccurate results.

This study aims to evaluate the efficiency of recently fine-tuned versions of BERT (Joshi et al., 2022; Joshi, 2022) for Marathi in extrinsic plagiarism detection– a method that identifies plagiarism by comparing input documents with a reference database of texts. We propose a system that integrates BERT embeddings with TF-IDF (Salton and Buckley, 1987) vectors, advancing the research and development of hybrid plagiarism detection models that combine syntactic, semantic, and statistical features of low-resource languages to achieve more accurate classifications.

The contributions[1] of this paper are as follows:

- Exploring the application of language models fine-tuned on Marathi to assess their efficiency in semantic analysis.

- Developing a plagiarism detection system that combines TF-IDF with BERT embeddings enhances feature extraction and the analysis of Marathi texts, leading to more accurate results for low-resource languages where features mined using fine-tuned language models may not suffice. This approach contributes significantly to content moderation, plagiarism detection, and paraphrase identification fields.

- Introducing a novel, ensemble-based method for semantic-based plagiarism detection specifically tailored for the Marathi language.

- Developing a labeled corpus for paraphrase and plagiarism detection using translation lan-

---

[1]The experiment code and dataset created can be found here: https://github.com/aditya-choudhary599/Marathi-Plagiarism-Detection

guage models to support and advance research in this area.

## 2 Previous Work

Shenoy and Potey (2016) and Naik et al. (2019) explored the use of WordNet (Miller, 1995) to capture semantic relations among Marathi words for plagiarism detection. In addition to WordNet, Shenoy and Potey (2016) employed lexical features such as n-grams (Shannon, 1948), syntactic features like Part-Of-Speech (POS), structural analysis, and Naive Bayes classification (Lewis, 1998) for detecting plagiarism. Meanwhile, Srivastava and Govilkar (2019) developed a paraphrase detection system that utilized Universal Networking Language (UNL) Graph-Based Similarity (Uchida et al., 2005) to measure semantic similarity, alongside metrics like Sumo Metric (Cordeiro et al., 2007), Jaccard (Jaccard, 1901), Cosine (Salton et al., 1975), and Word Order similarity for assessing statistical similarity in Marathi texts.

While Mahender and Solanke (2022) did utilize BERT to create word embeddings and compute cosine similarity between paraphrased Marathi words and sentences, their study focused solely on analyzing Levenshtein distances (Levenshtein, 1966) and cosine similarity without developing a classification model for identification. Lastly, C. Namrata Mahender, Ramesh Ram Naik (2020) and Kale and Prasad (2018) adopted a stylometry-based approach to identify plagiarized texts, using lexical features along with metrics like Hapax Legomena and Hapax DisLegomena to evaluate vocabulary richness.

The previous work on paraphrase and plagiarism detection in Marathi has not fully explored the efficiency of BERT for semantic-based extrinsic plagiarism detection. BERT embeddings are superior in capturing semantic relationships, offering context-sensitive, dense vector representations of words through deep learning (Devlin et al., 2018). Joshi et al. (2022) introduced MahaSBERT-STS, a specialized variant of the SBERT (Sentence-BERT) model (Reimers and Gurevych, 2019) trained on Natural Language Inference (NLI) and Semantic Textual Similarity (STS) datasets, making it well-suited for accurately capturing semantic similarity in Marathi texts and identifying plagiarism.

Research on plagiarism and paraphrase detection has expanded to other Indian languages. For instance, Kong et al. (2016) and Sarkar (2016a) employed similarity measures, including cosine similarity, Jaccard similarity, edit distance, and Dice distance, to train Gradient Boosting Tree (He et al., 2019) and Probabilistic Neural Network (Specht, 1990) classification models, respectively, for identifying paraphrased texts in Hindi, Punjabi, Malayalam, and Tamil. In a similar approach, Bhargava et al. (2016) and Saini and Verma (2018) computed normalized IDF scores and word overlap, demonstrating the high performance of Random Forest classifiers in their analyses. Additionally, Sarkar (2016b) utilized cosine similarity through TF-IDF vectorization, alongside word overlap and semantic similarity via Word2Vec (Mikolov et al., 2013), to train a multinomial logistic regression model aimed at identifying paraphrasing in Indian languages. Furthermore, Bhargava et al. (2017) proposed deep learning models based on Convolutional Neural Networks and Recurrent Neural Networks for paraphrase detection in both Hindi and English, assessing the effectiveness of WordNet and Word2Vec embeddings for feature extraction.

Previous works largely used precomputed similarity scores as input features to classification models, with many of these scores lacking semantic depth, which limited the models to learning from the scores rather than from the text itself. Additionally, Word2Vec embeddings, while useful, provide static representations of words, overlooking context–a limitation addressed by BERT embeddings, which adapt to the context of each word.

Studies in plagiarism and paraphrase detection have shown that combining statistical features, such as TF-IDF vectorization, with semantic features from deep learning models enhances detection performance. For instance, Arabi and Akbari (2022) integrated semantic features from WordNet and FastText (Joulin et al., 2016) with TF-IDF weighting for effective plagiarism detection. Similarly, Agarwal et al. (2018) combined CNN-LSTM (Shi et al., 2015) and WordNet-based semantic features with statistical measures like TF-IDF similarity and n-gram overlap to improve paraphrase detection. These studies underscore the potential of hybrid approaches, especially for low-resource languages, where features extracted from fine-tuned BERT models alone may not yield optimal results.

## 3 Methodology

Instead of relying on precomputed similarity scores and overlaps, our approach involves feeding the model with direct numeric representations of the text, enabling it to learn from the inherent patterns in the language rather than abstracted metrics. The following sections cover our data collection and preprocessing procedures, the method for text representation and feature extraction, the proposed system architecture and implementation details.

| No. | Reference | Input | Label |
|---|---|---|---|
| 1 | A boy is jumping on skateboard in the middle of a red bridge. | The boy does a skateboarding trick. | 1 |
| 2 | A boy is jumping on skateboard in the middle of a red bridge. | The boy skates down the sidewalk. | 0 |
| 3 | Two blond women are hugging one another. | There are women showing affection. | 1 |
| 4 | Two blond women are hugging one another. | The women are sleeping. | 0 |

Table 1: Samples from the Dataset. Here, label '1' indicates that input text was plagiarized or paraphrased from the reference text

| Model | Metric | Score |
|---|---|---|
| aryaumesh/ english-to-marathi | **BERT Precision** | **88.57%** |
| | **BERT Recall** | **88.60%** |
| | **BERT F1** | **88.58%** |
| | **TransQuest Score** | **0.72** |
| Helsinki-NLP/ opus-mt-en-mr | BERT Precision | 71.00% |
| | BERT Recall | 67.82% |
| | BERT F1 | 69.33% |
| | TransQuest Score | 0.60 |

Table 2: Comparison of translation models using BERTScore (precision, recall, F1) and TranQuest Score.

### 3.1 Data Collection

Previous work on Marathi text plagiarism and paraphrase detection has often lacked a standardized dataset, with many datasets being manually created or translated from other sources. To address this limitation, we constructed our dataset by translating the MIT Plagiarism Detection Dataset[2]. This dataset is a modified subset of the Stanford Natural Language Inference (SNLI) Corpus (Bowman et al., 2015), which is widely used for sentence similarity tasks. The SNLI corpus categorizes pairs of sentences into entailment, contradiction, or neutral, making it highly applicable for plagiarism and paraphrase detection.

The MIT Plagiarism Detection Dataset dataset contains 366,915 labeled pairs of reference and input short texts, with labels indicating the presence or absence of plagiarism. Table 1 illustrates a few sample pairs from the dataset.

We evaluated the BERTScores (Zhang* et al., 2020) and TransQuest scores (Ranasinghe et al., 2020b,a) achieved by the following models that we considered for translating the dataset:

- Helsinki-NLP/opus-mt-en-mr, developed by the Helsinki NLP group as part of the OPUS-MT project (Tiedemann et al., 2023; Tiedemann and Thottingal, 2020).

- The Google Translate API[3]. While this API produced accurate translations, its rate-limiting restricted our ability to use it for the complete dataset and was hence not used.

- aryaumesh/english-to-marathi[4], a fine-tuned Multilingual BART (mBART) model (Liu et al., 2020) trained with 611 million parameters for English-to-Marathi translation.

BERTScore calculates the precision, recall, and F1 scores for the translations, while the TranQuest score is a value between 0 and 1, where 1 indicates a perfect translation. We used the monotransquest-da-en_any[5] model, a sentence-level TransQuest architecture, for calculating the TransQuest score. Finally, we chose the aryaumesh/english-to-marathi model for translating the dataset due to its superior performance (as seen in table 2).

---

[2]https://www.kaggle.com/datasets/ruvelpereira/mit-plagairism-detection-dataset

[3]https://py-googletrans.readthedocs.io/en/latest/

[4]https://huggingface.co/aryaumesh/english-to-marathi

[5]https://huggingface.co/TransQuest/monotransquest-da-en_any

| No. | English Text | Marathi Translation |
|-----|--------------|---------------------|
| 1 | A person on a horse jumps over a broken down airplane. | घोड्यावर असलेला माणूस तुटलेल्या विमानावर उडी मारतो |
| 2 | A boy is jumping on skateboard in the middle of a red bridge. | लाल पुलाच्या मधोमध एक मुलगा स्केटबोर्डवर उडी मारत आहे. |
| 3 | A few people in a restaurant setting, one of them is drinking orange juice. | रेस्टॉरंटच्या सेटिंगमध्ये काही लोकं, त्यापैकी एक संत्रीचा रस पित आहे. |

Table 3: Translation Examples from our Generated Dataset

## 3.2 Data Preprocessing

To preprocess the data, we removed punctuation and stop words[6] from the text. Next, we applied rule-based suffix stripping for stemming and lemmatization to normalize the Marathi texts, ensuring consistent root forms. The cleaned and processed data was then prepared for feature extraction.

## 3.3 Text Representation and Feature Extraction

From the cleaned Marathi texts, we generated BERT embeddings and TF-IDF vectors for each pair of reference and input texts, considering each text as an individual document. For embeddings, we employed the `MahaSBERT-STS` model (Joshi et al., 2022), available through Hugging Face[7]. This model was chosen due to its specific training on Semantic Textual Similarity (STS) datasets, optimizing its effectiveness in capturing the semantic similarity of paraphrased or plagiarized Marathi texts. The `MahaSBERT-STS` model generates embeddings of dimension (768x1) for each sentence.

To evaluate model performance across various embedding dimensions, we also created reduced-dimensional embeddings at (512x1) and (256x1) using Principal Component Analysis (PCA) (Abdi and Williams, 2010). Additionally, we generated TF-IDF vectors of dimensions (256x1) and (400x1) to train our models on a range of vector representations.

Finally, we performed element-wise subtraction of the BERT embeddings of the input texts from those of the reference texts in their respective pairs, obtaining semantic vectors to represent the relationships between each pair of texts. The same element-wise subtraction process was applied to the TF-IDF vectors. Figure 1 illustrates the complete feature extraction pipeline for generating the BERT and TF-IDF vectors.

---

[6] https://github.com/stopwords-iso/stopwords-mr
[7] https://huggingface.co/l3cube-pune/marathi-sentence-similarity-sbert

We utilized 80% of the extracted vectors for training the classifier and reserved 20% for testing. To further validate the classifiers performance and assess the dataset's quality, we evaluated the model on the Microsoft Research Paraphrase Corpus (translated into Marathi using the same translator model described in Section 3.1). This step underscores the dataset's potential for training plagiarism detection models. Detailed results and comparisons are presented in Section 4.2.

## 3.4 Proposed System

The proposed system (Figure 2) employs a weighted ensemble approach (Dietterich, 2000), leveraging classifiers trained on distinct text representations– pairwise BERT embeddings (BERT classifiers) and TF-IDF vectors (TF-IDF classifiers). This ensemble method integrates the unique strengths of both text representations: while BERT embeddings capture semantic nuances (Devlin et al., 2018; Reimers and Gurevych, 2019) essential for detecting paraphrased and plagiarized text, TF-IDF vectors preserve statistical and syntactic information in Marathi text.

We evaluated multiple classification models, including Random Forest (Breiman, 2001), XGBoost (Chen and Guestrin, 2016), LightGBM (Ke et al., 2017), Support Vector Classifier (SVC) (Cortes, 1995), Decision Tree (Loh, 2011), Naive Bayes, AdaBoost (Freund and Schapire, 1997) and Logistic Regression (Cox, 1958), on BERT embeddings of dimensions (768x1), (512x1), and (256x1), as well as on TF-IDF embeddings of dimensions (400x1) and (256x1). For optimal model configurations, we tuned hyperparameters using FLAML (Wang et al., 2021) and GridSearchCV[8], record-ing the performance metrics specified in subsection 3.5.

---

[8] https://scikit-learn.org/dev/modules/generated/sklearn.model_selection.GridSearchCV.

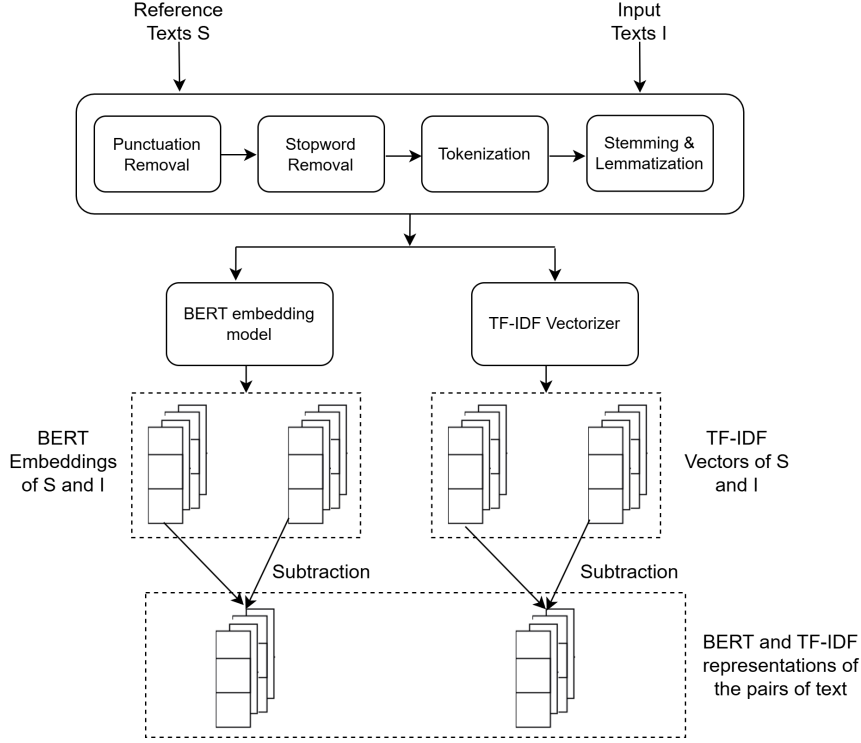

Figure 1: Pipeline for Extracting Features from Reference and Input Text Pairs

In the ensemble, each classifier predicts the probability of an input text being plagiarized from the reference text, based on the text representation it was trained on (BERT or TF-IDF). We calculate net probabilities for each classifier set as weighted averages:

$$P_{BERT} = \sum_{i=1}^{N1} p_{Bi} \cdot w_{Bi}$$

$$P_{TF-IDF} = \sum_{j=1}^{N2} p_{Tj} \cdot w_{Tj}$$

where $N1$ and $N2$ represent the number of BERT and TF-IDF classifiers, respectively; $p_{Bi}$ and $p_{Tj}$ are the probabilities predicted by each classifier in the BERT and TF-IDF sets, and $w_{Bi}$ and $w_{Tj}$ are the corresponding weights assigned to each classifier. The final probability $P$ is then computed as the weighted average of $P_{BERT}$ and $P_{TF-IDF}$:

$$P = P_{BERT} \cdot W_{BERT} + P_{TF-IDF} \cdot W_{TF-IDF}$$

where $W_{BERT}$ and $W_{TF-IDF}$ are the ensemble weights assigned to each set. The input text is classified as plagiarized if $P > 0.5$.

---

html

Model combinations and weights were iteratively refined to achieve optimal performance by leveraging complementary insights from each classifier set, as documented in subsection 4.1. The final system specifications are outlined in Table 7.

### 3.5 Evaluation Metrics

In this study, we evaluated the performance of our model using accuracy, precision, recall, and F1 score. Additionally, we analyzed the AUC score and examined the variance in these metrics as the weight assigned to BERT embeddings ($W_{BERT}$) was adjusted. This analysis provides insights into the influence of BERT embeddings on the final classification outcome and highlights the complementary role of TF-IDF-based text representations in the system.

## 4 Results and Discussion

We evaluated and fine-tuned various classification models, including our proposed weighted ensemble system, to achieve optimal performance. Table 4 presents the best results for each model based on both TF-IDF and BERT feature representations, detailed as follows.

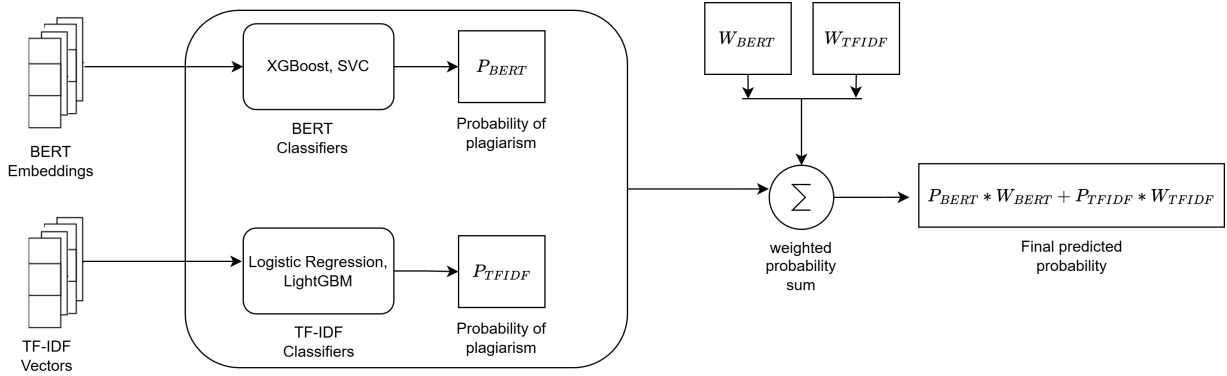

Figure 2: Proposed Weighted Ensemble Voting System for Plagiarism Detection

## 4.1 System Specifications

To maximize system performance, we experimented with different model combinations, weight distributions, and dimensions of TF-IDF and BERT text representations. Results indicated that Logistic Regression and LightGBM, assigned weights of 0.1 and 0.9, respectively, and trained on TF-IDF vectors of size 400, performed well when used in conjunction with XGBoost and SVC, weighted 0.7 and 0.3 and trained on BERT embeddings of size 768. The ensemble system achieved optimal results with $W_{BERT}$ and $W_{TF-IDF}$ values of 0.6 and 0.4, respectively.

This configuration demonstrated the advantage of integrating insights from both TF-IDF vectors and BERT embeddings, yielding more accurate results than models trained exclusively on BERT embeddings ($W_{BERT} = 1$) or TF-IDF vectors ($W_{TF-IDF} = 1$). The complete system specifications and hyperparameters for each classifier are detailed in Table 7.

## 4.2 Evaluation and Comparison

Our proposed system, utilizing models trained on both TF-IDF and BERT feature representations, achieved the highest accuracy of 82.04%, compared to 80.64% accuracy when using only BERT embeddings. This system demonstrated the highest accuracies across all data inputs, as shown in Table 4.

We observed that most models performed best with BERT embeddings of size 768, except for Logistic Regression, which yielded improved results on 256-sized embeddings. While individual models like Random Forest, XGBoost, and LightGBM achieved high scores, combining them in our ensemble system did not result in the highest accuracy. Logistic Regression and Decision Tree had lower standalone accuracy scores (65.67% and 64.87%, respectively) due to limitations in high-dimensional spaces, yet they contributed effectively within the ensemble system.

Some TF-IDF models displayed high recall rates which indicates a strong capacity for capturing true positives. However, these models also showed lower precision, reflecting a higher rate of false positives. This trade-off highlights TF-IDFs tendency to be more inclusive in its classifications, leading to a lower threshold for positive cases. When used alongside BERT embeddings, this strength in true positive identification proved beneficial for the ensemble system.

Table 5 demonstrates that our proposed system achieved superior performance scores on the validation data compared to previously used top-performing models. These results highlight not only the robustness of our system but also the applicability and quality of our translated dataset for plagiarism detection tasks.

## 4.3 Comparison with Previous Approach

Most previous approaches focused on computing various similarity measures between pairs of source and input texts, followed by training machine learning models on these measures to predict whether the input text was plagiarized. While this method is simpler to implement, it limits classifiers to rely solely on computed metrics, preventing them from learning directly from text patterns. Consequently, this leads to a loss of contextual information about semantic relationships, which is crucial for plagiarism detection. Moreover, such approaches often perform poorly for paraphrased texts, where surface-level similarity measures may yield low scores.

Table 6 compares the performance of our pro-

| Data | Model | Data Dimension | Accuracy | Precision | Recall | F1 Score |
|---|---|---|---|---|---|---|
| Combined | **Proposed System** ($W_{BERT} = 0.6$) | TF-IDF(400) and BERT(768) | **82.04%** | **80.22%** | **85.32%** | **82.69%** |
| BERT | **Proposed System** ($W_{BERT} = 1$) | **768** | **80.64%** | **78.85%** | **83.92%** | **81.31%** |
| | Naive Bayes | 768 | 74.65% | 71.38% | 82.47% | 76.52% |
| | Logistic Regression | 256 | 65.67% | 64.58% | 69.72% | 67.05% |
| | Decision Tree | 768 | 64.87% | 63.35% | 70.92% | 66.92% |
| | SVC | 768 | 66.07% | 64.11% | 73.31% | 68.40% |
| | Random Forest | 768 | 77.64% | 77.91% | 77.29% | 77.60% |
| | Adaboost | 768 | 74.05% | 72.16% | 78.49% | 75.19% |
| | Xgboost | 768 | 77.84% | 77.43% | 78.97% | 78.19% |
| | LightGBM | 768 | 79.44% | 78.33% | 81.75% | 80.00% |
| TF-IDF | **Proposed System** ($W_{BERT} = 0$) | **400** | **58.68%** | **58.24%** | **63.10%** | **60.57%** |
| | Naive Bayes | 256 | 51.70% | 51.61% | 57.37% | 54.34% |
| | Logistic Regression | 256 | 53.69% | 53.64% | 55.78% | 54.69% |
| | Decision Tree | 400 | 54.49% | 52.53% | 95.22% | 67.71% |
| | SVC | 256 | 54.09% | 53.96% | 56.97% | 55.43% |
| | Random Forest | 256 | 55.49% | 53.18% | 93.23% | 67.73% |
| | Adaboost | 400 | 55.89% | 53.49% | 91.63% | 67.55% |
| | Xgboost | 400 | 58.68% | 55.29% | 91.63% | 68.97% |
| | LightGBM | 256 | 54.89% | 53.01% | 87.65% | 66.07% |

Table 4: Performance of the Proposed System and various Classifiers with the Data they were trained on posed system, which is trained directly on TF-IDF and BERT vectors using the model specifications detailed in Table 7, with the traditional approach that employs classifiers trained on precomputed similarity metrics. These metrics include FastText word embedding similarity, N-gram overlap, Levenshtein distance, Fuzzy string similarity, Jaccard similarity, and Cosine similarity, calculated for each text pair in the dataset. As illustrated, our proposed system significantly outperforms the traditional approach, showcasing its robustness and superior capability in capturing the complexities and nuances of plagiarism detection.

### 4.4 Impact of BERT Embeddings on Performance

We analyzed the impact of $W_{BERT}$ on the proposed system's accuracy (Figure 3a), precision (Figure 3b), F1 score (Figure 3c), and AUC score (Figure 3d). The metrics reached their optimal values when $W_{BERT}$ was set to 0.6, indicating that while performance improved as BERT-based predictions were weighted more heavily, it was only to a certain extent.

Interestingly, variations in accuracy and F1 score were nearly identical, both peaking at $W_{BERT} = 0.6$, suggesting that precision and recall varied in proportion to accuracy. This behavior likely reflects the balanced nature of our dataset, which maintains an even distribution of false positives and false negatives. Likewise, both precision and AUC score peaked at $W_{BERT} = 0.6$, highlighting the models effectiveness at accurately identifying true positives and distinguishing between classes. Recall remained stable, peaking only at $W_{BERT} = 0.6$.

These results demonstrate that using both TF-IDF and BERT embeddings in conjunction enhances system accuracy for plagiarism detection, particularly in low-resource languages.

## 5 Conclusion

We proposed a weighted ensemble voting system that leverages both TF-IDF and BERT-based text representations to detect extrinsic plagiarism and paraphrasing in Marathi text. Our system not only outperformed individual classification models but also demonstrated the complementary value of using TF-IDF vectors alongside BERT embeddings, resulting in enhanced classification accuracy over BERT-only and TF-IDF-only models. By exploring various model combinations, weight configurations, and embedding dimensions, we identified an optimal configuration that achieved a remarkable accuracy of 82.04% using BERT embeddings of size 768 from `MahaSBERT-STS` alongside TF-IDF vectors of size 400, thereby surpassing the performance of other classification models.

This study highlights the effectiveness of combining statistical text vectorization methods, such as TF-IDF, with context-based embeddings like BERT to capture both statistical and semantic as-pects of Marathi texts. This approach proves particularly beneficial for low-resource languages like Marathi, which lack extensive datasets and robust, domain-specific embeddings. Our results underscore the potential of hybrid text representation methods in addressing the unique challenges presented by languages with limited computational resources and linguistic tools.

| Model | Data Dimension | Accuracy | Precision | Recall | F1 Score |
|---|---|---|---|---|---|
| **Proposed System** | **TF-IDF(400) and BERT (768)** | **78.20%** | **80.74%** | **92.39%** | **86.17%** |
| XGboost | BERT(768) | 71.20% | 73.17% | 98.68% | 84.03% |
| LightGBM | BERT(768) | 73.19% | 75.34% | 97.61% | 85.04% |
| Random Forest | BERT(768) | 70.59% | 72.32% | 94.53% | 81.95% |

Table 5: Performance of Classifiers on Validation Data

| Classifier | Accuracy (%) | Precision (%) | Recall (%) | F1 Score (%) |
|---|---|---|---|---|
| **Proposed System** | **82.04** | **80.22** | **85.32** | **82.69** |
| Random Forest | 69.67 | 68.54 | 69.79 | 69.15 |
| XGBoost | 68.26 | 68.04 | 69.34 | 68.69 |
| LightGBM | 70.17 | 69.81 | 70.45 | 70.13 |
| Naive Bayes | 64.32 | 63.45 | 65.21 | 64.32 |

Table 6: Performance Comparison of Proposed System (trained on TF-IDF and BERT vectors) with Previous Approach (classifiers trained on pre-computed similarity measures)

In conclusion, our system presents a promising, adaptable solution for accurate and efficient plagiarism and paraphrase detection in Marathi. The adaptability of our approach suggests it could be extended to similar low-resource languages, potentially facilitating more robust and inclusive text analysis tools across diverse linguistic contexts. This work paves the way for further exploration into optimized ensemble systems that can harness the strengths of both traditional and advanced text representation methods.

## Limitations

This study contributes to advancing plagiarism detection for the Marathi language by leveraging language models like BERT and statistical vectorizers like TF-IDF. However, some limitations should be noted.

First, the absence of standardized, well-annotated datasets for Marathi posed challenges in benchmarking our model effectively against existing systems.

Further, the limited availability of a large corpus for fine-tuning Marathi-specific BERT models, in contrast to widely resourced languages like English, may have impacted performance. Access to BERT models trained on a more extensive Marathi corpus could better address the unique linguistic characteristics of Marathi, potentially improving the capture of semantic nuances and contextual relationships.

Also, the dataset used for training primarily consists of short-text pairs, which makes the approach effective for detecting paraphrased and semantically modified plagiarism. However, its applicability to longer academic texts or creative works remains untested. Future research should explore adaptations such as segmenting lengthy academic texts into smaller coherent chunks or incorporating stylometric analysis for creative writing.

Lastly, limited computing power and GPU resources extended training times and restricted the scope of experimentation to determine optimal system parameters.

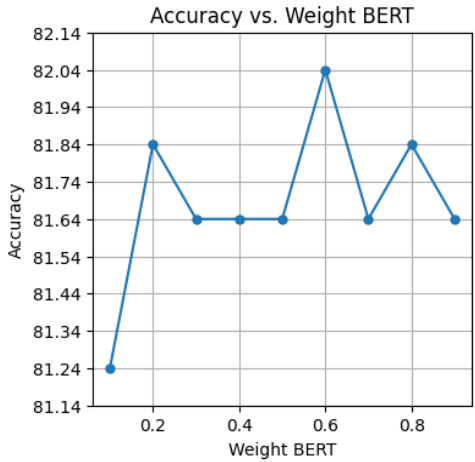

(a) Accuracy v/s $W_{BERT}$ for combined system

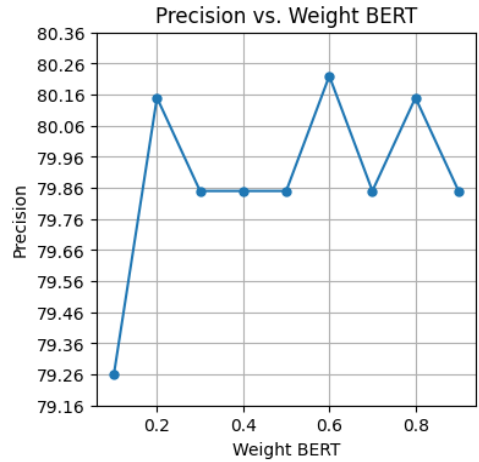

(b) Precision v/s $W_{BERT}$ for combined system

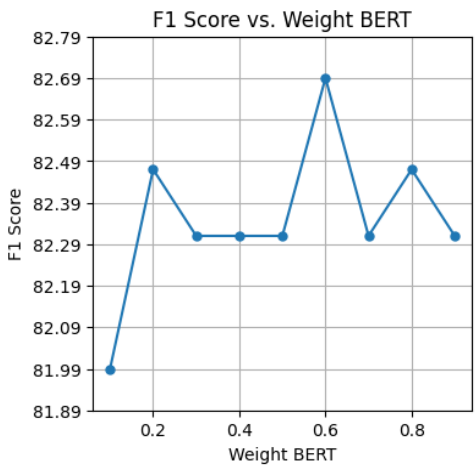

(c) F1 Score v/s $W_{BERT}$ for combined system

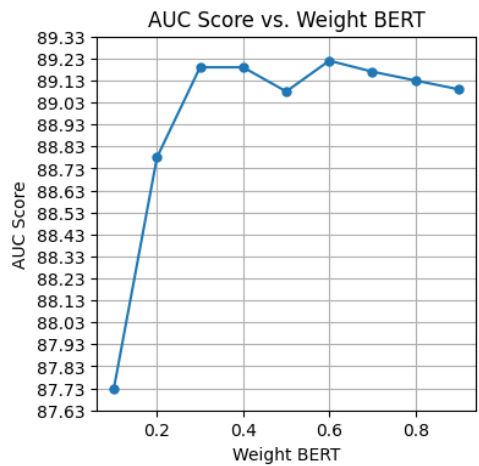

(d) AUC Score v/s $W_{BERT}$ for combined system

# A  Appendix

| TF-IDF Classifiers | | | | BERT Classifiers | | | |
|---|---|---|---|---|---|---|---|
| $W_{TF-iDF} = 0.4$ | | | | $W_{BERT} = 0.6$ | | | |
| **Classifier** | **Hyperparam** | **Value** | $w_{Tj}$ | **Classifier** | **Hyperparam** | **Value** | $w_{Bi}$ |
| Logistic Regression | C | 0.136 | 0.1 | XGBoost | colsample_bylevel | 0.198 | 0.7 |
| | penalty | l2 | | | colsample_bytree | 0.444 | |
| LightGBM | colsample_bytree | 0.929 | 0.9 | | grow_policy | lossguide | |
| | learning_rate | 0.185 | | | learning_rate | 0.165 | |
| | max_bin | 15 | | | max_leaves | 20 | |
| | min_child_samples | 12 | | | min_child_weight | 0.270 | |
| | n_estimators | 1 | | | n_estimators | 371 | |
| | num_leaves | 8 | | SVC | kernel | rbf | 0.3 |
| | reg_alpha | 0.002 | | | C | 100 | |
| | reg_lambda | 0.159 | | | degree | 2 | |
| | | | | | gamma | scaler | |
| | | | | | max_iter | 1000 | |

Table 7: Proposed System Specifications